\documentclass[runningheads]{llncs}
\usepackage[T1]{fontenc}
\usepackage{graphicx}
\usepackage{xspace}
\usepackage[numbers]{natbib}
\usepackage{booktabs}
\usepackage{makecell}
\usepackage{bbding}
\usepackage{url}

\usepackage{multirow}
\usepackage{amsmath}
\usepackage{amssymb}
\usepackage[most]{tcolorbox}
\usepackage[utf8]{inputenc}
\usepackage{subcaption}

\newcommand{\agentname}{\textsc{RAL-Writer}\xspace}
\newcommand{\yes}{\Checkmark}
\newcommand{\noh}{\XSolidBrush}

\begin{document}

\title{Lost-in-the-Middle in Long-Text Generation: Synthetic Dataset, Evaluation Framework, and Mitigation Paradigm}

\titlerunning{Lost-in-the-Middle in Long-Text Generation}

\author{Junhao~Zhang\inst{1,2} \and
    Richong~Zhang\inst{1}\and
    Fanshuang~Kong\inst{1,3}\Envelope \and
    Ziyang~Miao\inst{1} \and
    Yanhan~Ye\inst{1} \and
    Yaowei~Zheng\inst{1}}

\authorrunning{J. Zhang et al.}

\institute{
    School of Computer Science and Engineering, Beihang University, Beijing, China
    \and
    National College for Excellent Engineers, Beihang University, Beijing, China
    \and
    School of Software, Beihang University, Beijing, China
    \email{
        \{zhangjunhao,zhangrichong,kongfs,miaozy,yeyanhan,hiyouga\}@buaa.edu.cn
    }
}

\maketitle

\begin{abstract}
    Existing long-text generation methods produce lengthy outputs from short inputs, leaving long-input-to-long-output generation underexplored. As input length increases, LLMs increasingly overlook information in the middle of the context---a limitation known as the ``lost-in-the-middle'' phenomenon---leading to inconsistent and incoherent outputs. To address this problem, we propose \textbf{R}etrieval-\textbf{A}ugmented \textbf{L}ong-Text \textbf{Writer} (\agentname), a training-free framework consisting of a \textit{Planner} that generates writing steps and a \textit{Writer} that produces content according to these steps. The \textit{Writer} jointly models semantic relevance and positional bias to compute importance scores, dynamically retrieve critical input segments, and strategically restate them. We also construct a benchmark dataset for long-input-to-long-output generation and introduce three evaluation metrics covering length, consistency, and quality. We evaluate \agentname against comparable baselines on this dataset. The results demonstrate the effectiveness of our approach. Code is available at \url{https://github.com/OnlyAR/RAL-Writer}.

\end{abstract}

\section{Introduction}

\begin{figure}[t]
    \centering
    \includegraphics[width=0.9\linewidth]{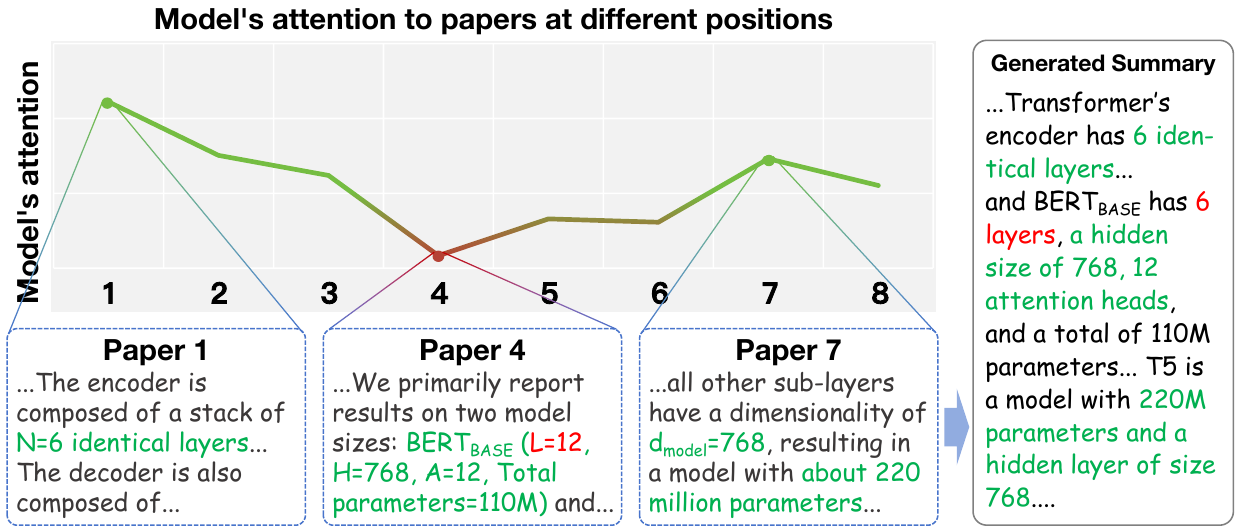}
    \caption{Our empirical results show that LLMs are more error-prone when using information from the middle of long inputs (\S\ref{subsec:lost}).}
    \label{fig:example}
\end{figure}

Although long-context LLMs can process extensive inputs, they often fail to satisfy output-length instructions: when asked to produce a 10,000-word paper, they typically generate fewer than 2,000 words~\cite{bai2024longwriter}. This limitation, commonly addressed through multi-step agents~\citep{xiong2025beyond}, long-response SFT~\citep{bai2024longwriter}, or preference alignment~\citep{pham-etal-2024-suri}, partly stems from limited exposure to long-form outputs during supervised fine-tuning. Yet real-world applications--including report generation from extensive logs, document-level summarization, and article continuation--require long-form generation conditioned on long inputs. This long-input-to-long-output setting remains underexplored and poses challenges to both long-context understanding and sustained generation.

Long inputs exacerbate position-dependent under-utilization in LLMs~\citep{liu-etal-2024-lost}. This ``lost-in-the-middle'' phenomenon causes models to overlook essential information at intermediate positions. In realistic long-document scenarios, this degradation can propagate to generation and hinder source-consistent writing (Figure~\ref{fig:example}). Existing benchmarks commonly assess long-context understanding and long-form generation separately (Table~\ref{tab:benchmarks-comparison}), leaving limited support for evaluating how models integrate information from different input positions into a coherent long-form output.

We therefore systematically study how position-dependent under-utilization affects long-form generation and evaluate the \textbf{R}etrieval-\textbf{A}ugmented \textbf{L}ong-Text \textbf{Writer} (\agentname), a practical mitigation built on a plan-and-write workflow. \agentname comprises a \textit{Writing Step Planner} and a \textit{Retrieve-and-Restate Writer} and requires neither additional model training nor architectural changes to the backbone. The \textit{Planner} generates writing steps for a long output based on the source input, and the \textit{Writer} produces the content step by step. During writing, \agentname combines step-specific semantic relevance with positional compensation to retrieve potentially under-utilized source chunks and restates them at the end of the prompt.

\begin{table}[t]
    \centering
    \small

    \resizebox{0.9\columnwidth}{!}{
        \setlength{\tabcolsep}{2pt}
        \begin{tabular}{cccccc}
            \toprule
            \textbf{Method}                                                  & \makecell{\textbf{Long Input}\\ \textbf{(> 8K)}} & \makecell{\textbf{Long Output}\\ \textbf{(> 1K)}} & \makecell{\textbf{Real-world}\\ \textbf{Aligned}} & \makecell{\textbf{Consistency}\\ \textbf{Evaluation}} & \makecell{\textbf{Quality}\\ \textbf{Evaluation}} \\
            \midrule
            \makecell{NIAH~\citep{kamradt2024niah}}                          & \yes                                             & \noh                                              & \noh                                              & \yes                                                  & \noh                                              \\
            \makecell{RULER~\citep{hsieh2024ruler}}                          & \yes                                             & \noh                                              & \noh                                              & \yes                                                  & \noh                                              \\
            \makecell{$\infty$Bench~\citep{zhang-etal-2024-bench}}           & \yes                                             & \noh                                              & \yes                                              & \yes                                                  & \noh                                              \\
            \makecell{SummHay~\citep{laban-etal-2024-summary}}               & \yes                                             & \noh                                              & \yes                                              & \yes                                                  & \noh                                              \\
            \midrule
            \makecell{LongGenBench$_{1}$~\citep{wu2024longgenbench}}         & \noh                                             & \yes                                              & \noh                                              & \noh                                                  & \noh                                              \\
            \makecell{LongGenBench$_{2}$~\citep{liu-etal-2024-longgenbench}} & \noh                                             & \yes                                              & \noh                                              & \noh                                                  & \noh                                              \\
            \makecell{LongWriter~\citep{bai2024longwriter}}                  & \noh                                             & \yes                                              & \yes                                              & \noh                                                  & \yes                                              \\
            \makecell{ProxyQA~\citep{tan-etal-2024-proxyqa}}                 & \noh                                             & \yes                                              & \yes                                              & \yes                                                  & \noh                                              \\
            \midrule
            Ours                                                             & \yes                                             & \yes                                              & \yes                                              & \yes                                                  & \yes                                              \\
            \bottomrule
        \end{tabular}
    }
    \caption{Recent representative benchmarks for long-context LLM evaluation. ``Real-world Aligned'' indicates whether tasks reflect practical applications, ``Consistency Evaluation'' assesses factual correctness, and ``Quality Evaluation'' measures language fluency and structural coherence.}
    \label{tab:benchmarks-comparison}
\end{table}

To evaluate this problem in a concrete, source-grounded setting, we construct a benchmark for long-input-to-long-output multi-paper scientific summarization. Models are required to generate comprehensive summaries of a specified length from multiple related scientific papers. This task naturally depends on long inputs, involves knowledge-intensive content, and offers practical value. The benchmark comprises 100 samples (300 papers collected from arXiv\footnote{\url{https://arxiv.org/}}) spanning diverse categories and document lengths.

In addition, we develop an evaluation framework covering three aspects: length, consistency, and quality. Length measures whether summaries satisfy specified requirements. Consistency uses carefully constructed question-answer (QA) pairs, generated and evaluated using advanced LLMs, to verify whether critical information across source papers is captured. Finally, quality--including fluency, coherence, and structural organization--is measured by a checklist-guided LLM judge. Together, these metrics go beyond surface-level measures such as ROUGE or BLEU to provide a task-specific, multifaceted assessment of long-text summarization. Our contributions are summarized as follows:

\begin{itemize}
    \item We systematically study how the ``lost-in-the-middle'' phenomenon manifests as position-dependent source under-utilization in long-form generation. Building on plan-and-write generation, we evaluate \agentname, a training-free retrieve-and-restate mitigation that provides step-specific reminders of relevant, potentially overlooked evidence.

    \item We construct a benchmark for long-input-to-long-output multi-paper scientific summarization and design a three-dimensional evaluation framework covering length adherence, source consistency, and writing quality.

    \item Experiments with two Qwen backbones show that \agentname improves overall consistency and quality over AgentWrite in this studied setting. Length adherence is mixed, and 16k-word generation remains challenging.
\end{itemize}

\begin{figure*}[!t]
    \centering
    \includegraphics[width=0.9\linewidth]{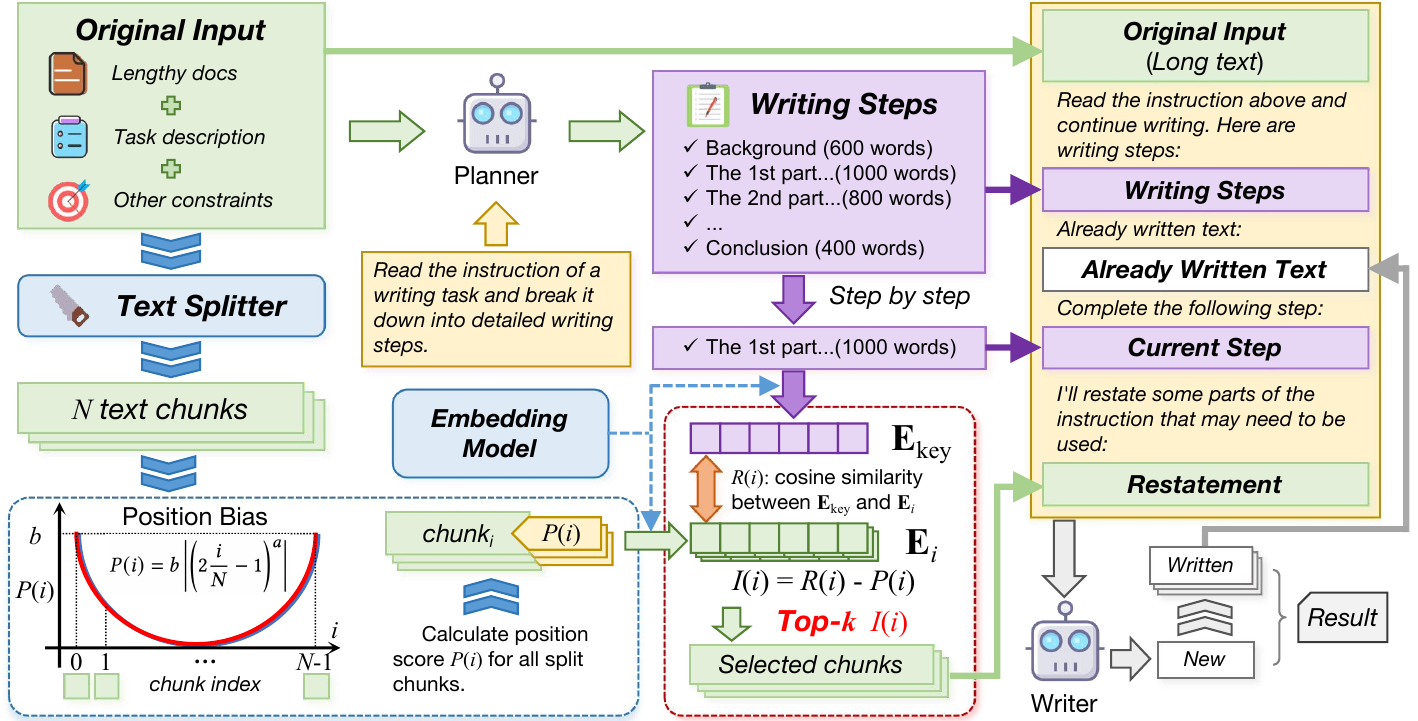}
    \caption{Illustration of \agentname.
        The \textit{Planner} decomposes the task into Writing Steps. The Text Splitter divides the input into chunks, and a relative-position score and a relevance score are computed for each chunk. At each writing step, the top-$k$ chunks with the highest importance scores are duplicated at the prompt tail as auxiliary reminders while the full original input remains available to the \textit{Writer}, which generates text step by step following the planned writing structure.
    }
    \label{fig:model}
\end{figure*}

\section{Related Work}

\textbf{Long-context and long-form generation.}
Extending the context window does not guarantee that LLMs can effectively use all available information. Existing work improves long-context processing through position-encoding and context-extension methods~\citep{peng2023yarn,han2024lm}. Other studies address long-form generation through long-response training and alignment~\citep{bai2024longwriter,pham-etal-2024-suri,ping2025longdpo} or iterative agent frameworks~\citep{xiong2025beyond,quan2024language}. However, these generation methods primarily target short instructions rather than long inputs that require information to be integrated throughout the generation process.

\textbf{Lost-in-the-middle mitigation.}
LLMs often underutilize information located in the middle of long contexts~\citep{liu-etal-2024-lost}. Prior work mitigates this issue through dense QA training~\citep{an2024make}, position-agnostic multi-step QA~\citep{he-etal-2024-never}, and position-encoding optimization~\citep{wuefficient}. These approaches focus mainly on comprehension or question answering. In contrast, we study how positional degradation affects long-form generation and mitigate it by retrieving and restating important information located in the middle of the context during writing.

\section{\agentname}

AgentWrite~\citep{bai2024longwriter} enables long-form generation via a ``Plan and Write'' paradigm but primarily targets short inputs. With long inputs, it suffers from the ``lost-in-the-middle'' problem (\S\ref{subsec:lost}). We therefore propose \agentname, comprising a \textit{Planner} and a content-generating \textit{Retrieve-and-Restate Writer}. Figure~\ref{fig:model} shows the workflow.

\subsection{Writing-Step Planner}
An effective strategy for generating long-form content with LLMs is to decompose the task and handle each part separately. The \textit{Planner} plays a crucial role in this decomposition process. It is designed to generate the overall structure and writing steps after comprehending lengthy inputs. Typically, the input consists of a long text and an instruction specifying the desired output length. After processing the entire input, the \textit{Planner} produces a comprehensive writing plan comprising multiple steps, with each step including specific writing requirements and an expected length. The total length of all steps is expected to precisely match the target length specified by the user.

\subsection{Retrieve-and-Restate Writer}
The \textit{Writer} follows the plan step by step, prompted with the input, prior text, and current requirements. To preserve important information, a retrieve-and-restate mechanism retrieves crucial chunks from the long text and strategically restates them.

\subsubsection{Long-text Chunking.} To ensure contextual coherence, we use a recursive splitter following LangChain~\citep{Chase_LangChain_2022}. We split by headings, paragraphs, tables, and sentences, and merge adjacent segments into overlapping chunks of at most 512 tokens. Following structural boundaries helps keep related content together, while overlap helps retain evidence that spans adjacent chunk boundaries. Each chunk is separately encoded with BGE-base-en-v1.5 and matched to the current writing step.

\begin{figure}[t]
    \centering
    \captionsetup[subfigure]{font=scriptsize}
    \begin{minipage}[t]{0.30\columnwidth}
        \begin{subfigure}[t]{\columnwidth}
            \centering
            \includegraphics[width=\columnwidth]{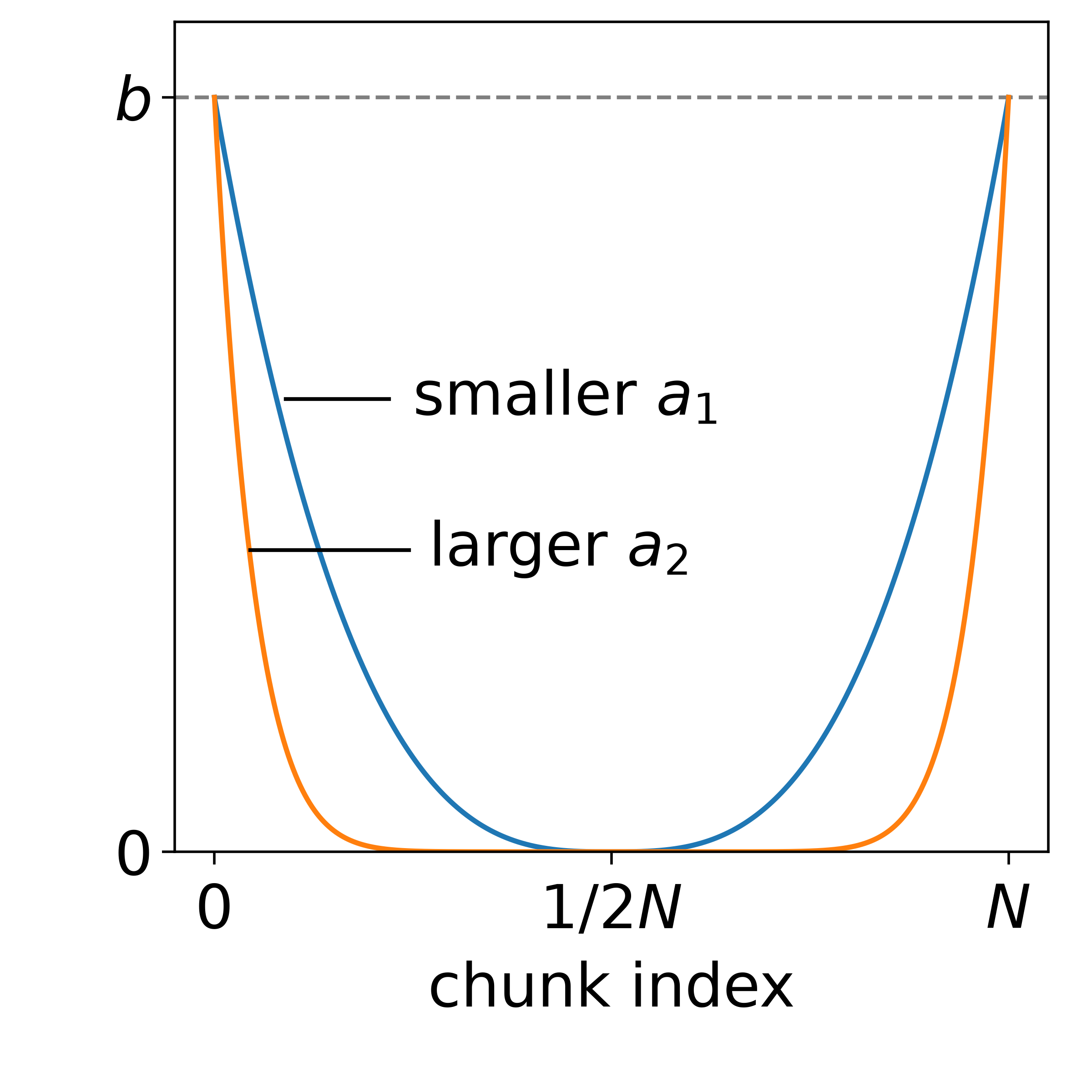}
            \caption{$P$ varies with $a$.}
            \label{fig:a-plot}
        \end{subfigure}
    \end{minipage}
    \begin{minipage}[t]{0.30\columnwidth}
        \begin{subfigure}[t]{\columnwidth}
            \centering
            \includegraphics[width=\columnwidth]{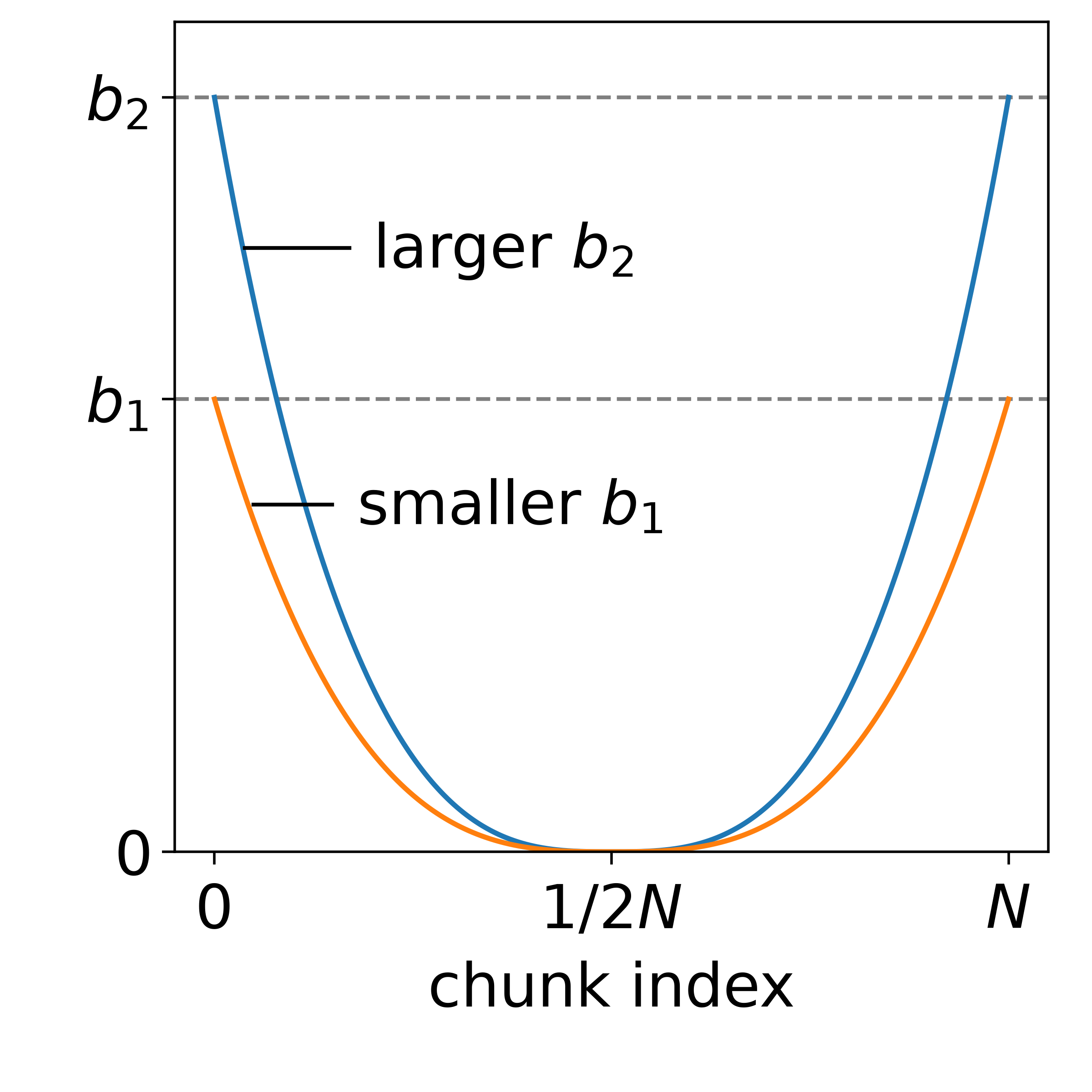}
            \caption{$P$ varies with $b$.}
            \label{fig:b-plot}
        \end{subfigure}
    \end{minipage}
    \begin{minipage}[t]{0.30\columnwidth}
        \begin{subfigure}[t]{\columnwidth}
            \centering
            \includegraphics[width=\columnwidth]{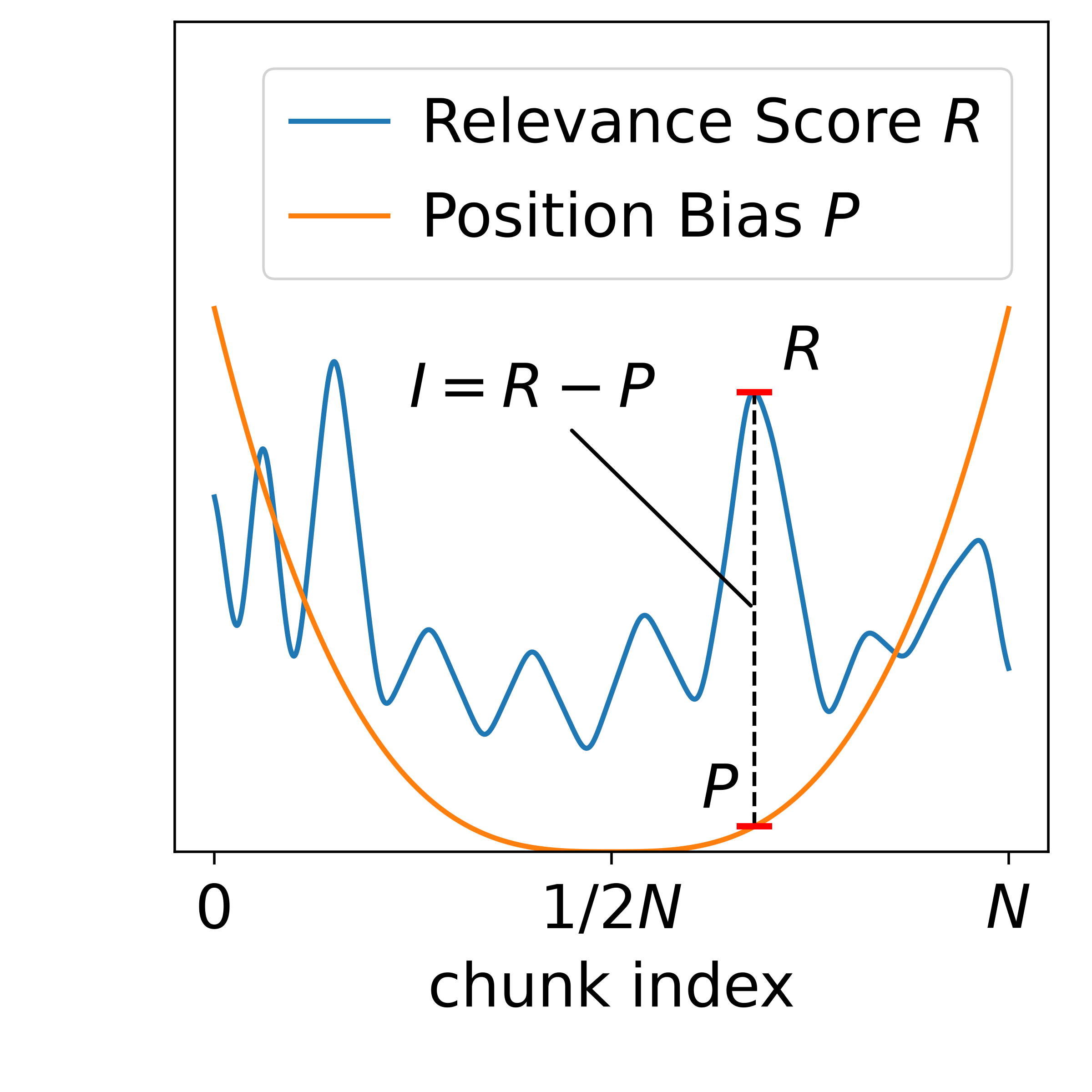}
            \caption{Schematic of $R$, $P$, and $I$.}
            \label{fig:I-plot}
        \end{subfigure}
    \end{minipage}
    \caption{Effects of $a$ and $b$ on position bias $P$ and importance $I$. (a) Larger $a$ steepens $P$ at the ends and flattens its center. (b) Larger $b$ raises the endpoint maxima. (c) $I=R-P$ favors relevant chunks near the context center.}
    \label{fig:P-R-I-plot}
\end{figure}

\subsubsection{Important-Chunk Retrieval.} Long inputs are susceptible to the ``lost-in-the-middle'' issue~\cite{liu-etal-2024-lost}. To direct the model toward source content that is relevant to the current writing step but may be under-utilized because of its position, we use an important-chunk retrieval mechanism. Intuitively, the mechanism prioritizes step-relevant chunks while applying a compensatory preference to those near the context middle. Let $\mathbf{E}_{i}$ and $\mathbf{E}_{\mathrm{key}}$ denote the embeddings of the $i$-th chunk and the current step, respectively. We define the Relevance Score $R(i)$ as the step-conditioned semantic relevance:
\begin{equation}
    R(i)=\frac{\langle \mathbf{E}_{i}, \mathbf{E}_{\mathrm{key}} \rangle}{||\mathbf{E}_{i}||\,||\mathbf{E}_{\mathrm{key}}||}.
\end{equation}

We introduce Relative Position Bias to capture each chunk's position in the context. Because chunks near the input ends receive more attention than those in the middle, the bias peaks at both ends, reaches its minimum at the center, and is symmetric about the midpoint. Formally:
\begin{equation}
    f(x) = b|2x - 1|^{a},
\end{equation}
where $x \in [0,1]$ is the normalized position and $a,b>0$ are tunable. The transformation $(2x-1)$ ensures symmetry, the absolute value sets minima at the center and maxima at the edges, and $a$ controls the steepness while $b$ sets the maximum score. For example, when $a=1$, the function decreases linearly toward the center, while larger values of $a$ yield a sharper drop, emphasizing the severity of the ``lost-in-the-middle'' effect.

To apply this formulation to discrete chunks, we consider an input divided into $N$ chunks. The relative position of the $i$-th chunk can be written as $x = i/N$. Substituting this into the function, the Position Bias becomes:
\begin{equation}
    P(i)
    =f\left(\frac{i}{N}\right)
    =b\left|2 \frac{i}{N} - 1\right|^{a}.
    \label{eq:P_i}
\end{equation}
Here, $R(i)$ captures relevance only to the current writing step; it is not an estimate of semantic density, novelty, or global discourse importance. In contrast, $P(i)$ is a tunable, symmetric prior for the positional advantage already enjoyed by boundary chunks, rather than another semantic score. It serves only as a compensatory prior and may be imperfect for documents with asymmetric structures or evidence distributions.

We use $P(i)$ as a residual correction that discounts the positional advantage of boundary chunks from $R(i)$, defining the Importance Score $I$ as:
\begin{equation}
    \begin{aligned}
        I(i) & = R(i)-P(i).
    \end{aligned}\label{eq:attention}
\end{equation}
For chunks with equal relevance, the lower $P(i)$ near the middle yields a higher $I(i)$; for chunks at the same position, subtracting the same positional term preserves their relevance ordering. Because $P(i)$ contains the scale $b$, the score is equivalently a weighted addition of a negative position prior to semantic relevance, with $b$ controlling the compensation strength. We use this residual form because it is interpretable and training-free, not because it is theoretically optimal; our experiments do not directly compare alternative score-fusion or reordering strategies. We retrieve the top-$k$ chunks ranked by $I$ for restatement. Figure~\ref{fig:P-R-I-plot} visualizes these scores.

\subsubsection{Restatement of Retrieved Chunks.}
At each writing step, the full original context remains available, and retrieved chunks are not removed from their original positions. Instead, the selected top-$k$ chunks are duplicated at the prompt tail as step-specific reminders. Within this appended block, we sort chunks in ascending order of their Importance Scores, placing higher-scoring chunks later in the prompt. This restatement changes the presentation rather than the source content, but duplication may introduce discourse discontinuity or ambiguity.

\section{Dataset Construction}

To assess long-text understanding and generation, we construct a dataset for generating structurally constrained summaries from multiple related scientific papers. Multi-paper summarization provides naturally long, knowledge-intensive inputs and requires coherent integration of information across documents.

Each sample comprises a long input formed by concatenating multiple scientific papers on a shared topic, often spanning tens of thousands of words. Models generate long-form summaries from these inputs, assessing their ability to integrate and distill knowledge from lengthy texts into narratives.

\begin{figure}[t]
    \centering
    \begin{tabular}{@{}m{0.40\columnwidth}<{\centering}@{\hspace{1mm}}m{0.40\columnwidth}<{\centering}@{}}
        \includegraphics[width=0.28\columnwidth]{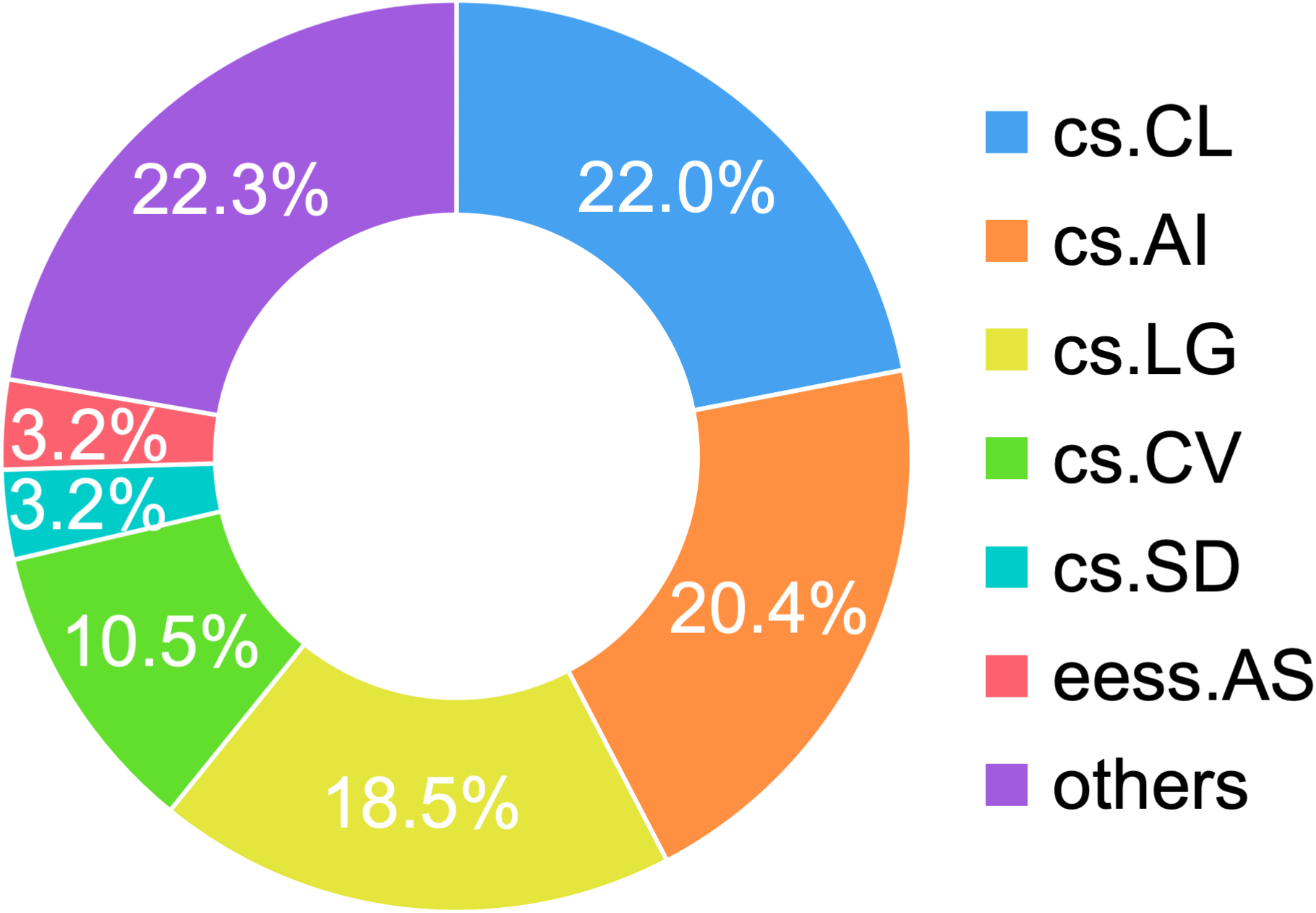} & \includegraphics[width=0.28\columnwidth]{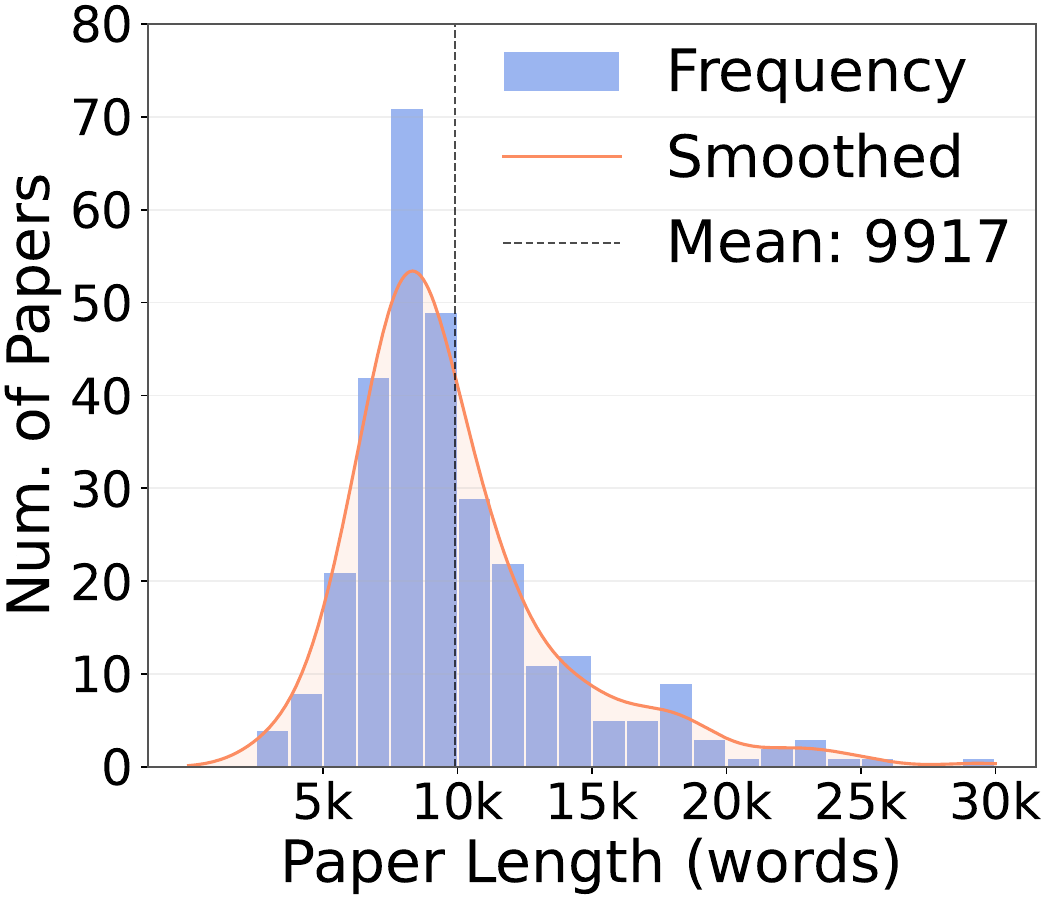} \\
        \small{(a) Category distribution}                                      & \small{(b) Length distribution}
    \end{tabular}
    \caption {Statistics of our dataset. Note that categories in (a) with proportions below 3\% are grouped into ``Others''.}
    \label{fig:papers-statistics}
\end{figure}

For each sample in the dataset, we collect a triplet of thematically related papers from arXiv, typically those that are introduced within the same survey or compared in the same study.
We download the corresponding \LaTeX\ source files and preprocess the data by removing extraneous elements such as comments, preambles, and appendices. To facilitate structural comprehension by large language models, we retain essential \LaTeX\ markup in the cleaned text. We further exclude papers with inconsistent formatting or insufficient length to ensure data quality. The resulting dataset consists of 100 samples (300 papers in total). Figure~\ref{fig:papers-statistics} summarizes key characteristics of the curated dataset, including the distribution of arXiv categories and document lengths.

\begin{figure}[t]
    \centering
    \includegraphics[width=0.9\columnwidth]{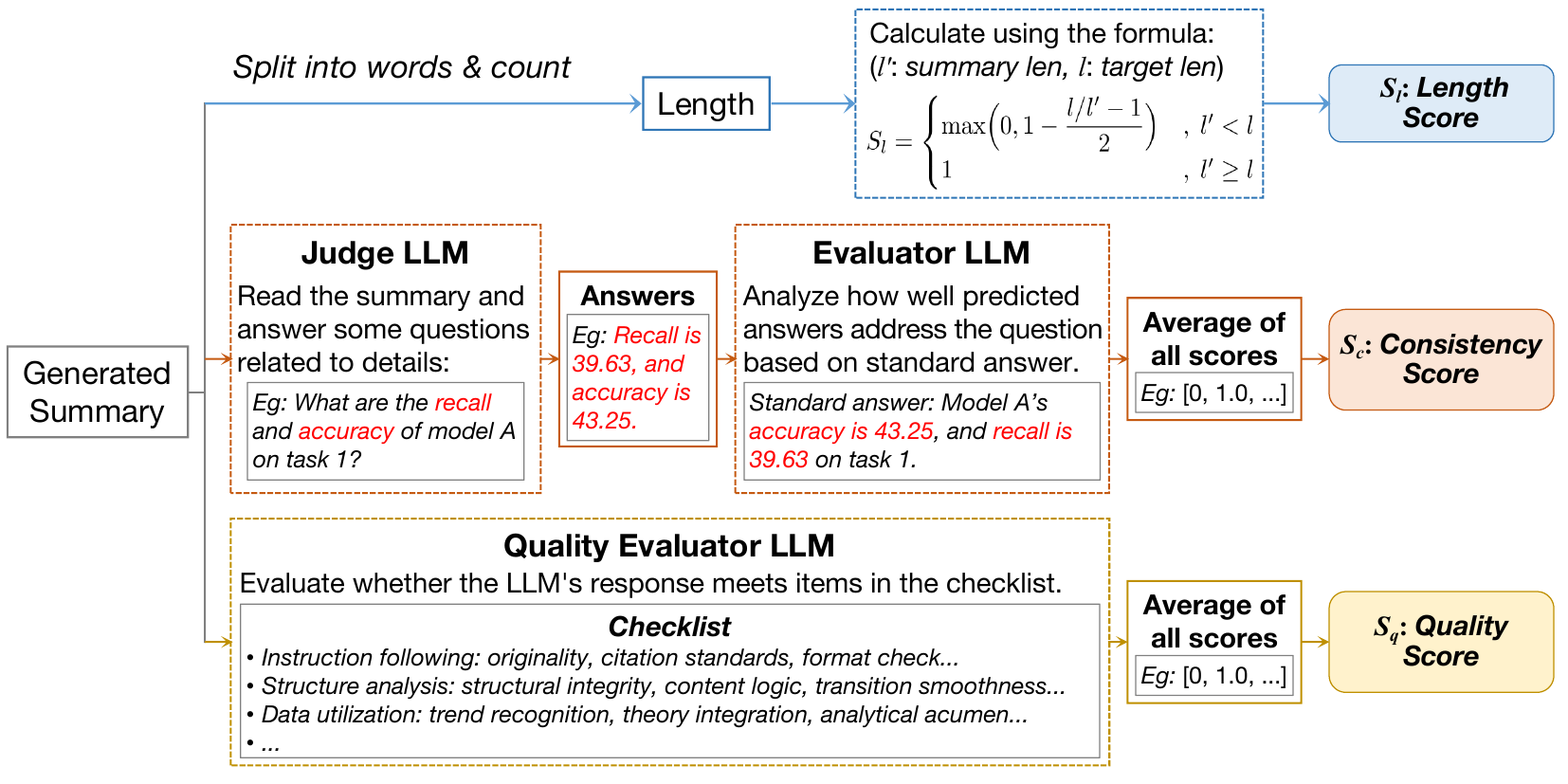}
    \caption{Overview of our evaluation framework.}
    \label{fig:benchmark-overview}
\end{figure}

\section{Evaluation}

To assess the generated summaries, we introduce three complementary metrics: length score, consistency score, and quality score.
These metrics are designed to capture different aspects of long-text summarization performance, ensuring a comprehensive evaluation.
Figure~\ref{fig:benchmark-overview} provides an overview of the task workflow and the evaluation framework.

\subsection{Length Sufficiency Evaluation}

Following LongBench-Write~\citep{bai2024longwriter}, we define the length sufficiency score as:
\begin{equation}
    S_l = \begin{cases}
        \max\Big(0, 1 - \dfrac{l/l' -1}{2}\Big) & , l' < l    \\
        1                                       & , l' \geq l
    \end{cases}\label{eq:length-score}
\end{equation}
where $l'$ and $l$ denote the generated and required lengths, respectively.

\subsection{Consistency Evaluation}

Following ProxyQA~\citep{tan-etal-2024-proxyqa}, we construct \textit{Single-Context} questions answerable from one paper and \textit{Cross-Context} questions requiring cross-paper synthesis. GPT-4o~\citep{hurst2024gpt} generates candidate QA pairs, which are filtered by an LLM and then manually selected and deduplicated. Each sample contains six questions of each type, totaling 1,200 QA pairs. A judge LLM answers each question from the generated summary, and an evaluator LLM scores the answer against the reference on a 0--1 scale~\citep{NEURIPS2023_91f18a12}:
\begin{equation}
    S_c = \frac{\bar{S}_\textrm{single} + \bar{S}_\textrm{cross}}{2},
\end{equation}
where $\bar{S}_\textrm{single}$ and $\bar{S}_\textrm{cross}$ are the mean scores for the two question types.

\noindent\textbf{Preliminary human validation.} We manually conducted a blinded pairwise assessment on a stratified subset of 48 QA instances from 12 topic clusters, evenly split between \textit{Single-Context} and \textit{Cross-Context} questions. For each instance, the AgentWrite and \agentname\ answers were anonymized and compared against the reference answer without revealing method identities or LLM scores. The human preference agreed with the LLM evaluator on 70.8\% of the answer pairs, with Kendall's $\tau_b=0.704$; after excluding instances in which either judgment was a tie, all 20 remaining directional comparisons agreed. This small QA-level study provides preliminary evidence for the evaluator's pairwise ranking behavior, but it does not validate full-summary factual consistency or the quality score $S_q$.

\subsection{Quality Evaluation}

We evaluate fluency, coherence, and structure with an LLM judge using a HelloBench-inspired checklist~\citep{que2024hellobench} of $C=8$ aspects and $N=5$ items per aspect:
\begin{equation}
    S_q = \frac{1}{C}\sum\limits_{i=1}^{C}\left(\frac{1}{N}\sum_{j=1}^{N}S_{i, j}\right),
\end{equation}
where $S_{i,j}$ is the score of checklist item $j$ under aspect $i$.

\section{Experiments}

\subsection{Experimental Setting}
We evaluate three open-source backbones—Qwen2.5-14B-Instruct, Qwen2.5-32B-Instruct~\citep{yang2024qwen2}, and LongWriter-glm4-9b~\citep{bai2024longwriter}—using vLLM~\citep{kwon2023efficient} on NVIDIA A100 GPUs, each with 40\,GB of memory. In \agentname, we use BGE-base-en-v1.5~\citep{bge_embedding} as the embedding model. For cost efficiency, GPT-4o-mini is used for question answering and answer evaluation, with temperatures of 0.3 and 0.

\begin{table*}[t]
    \centering
    \small
    \resizebox{\linewidth}{!}{
        \begin{tabular}{cc|ccc|ccc|ccc|ccc}
            \toprule
            \multirow{2}*{\textbf{Backbone}} & \multirow{2}*{\textbf{Method}} & \multicolumn{3}{c|}{\textbf{Overall}} & \multicolumn{3}{c|}{\textbf{4k}} & \multicolumn{3}{c|}{\textbf{8k}} & \multicolumn{3}{c}{\textbf{16k}}\\
            ~ & ~ & $S_l$ & $S_c$ & $S_q$ & $S_l$ & $S_c$ & $S_q$ & $S_l$ & $S_c$ & $S_q$ & $S_l$ & $S_c$ & $S_q$ \\
            \midrule
            GPT-4o & Single 
            & 0.00 & 23.14 & 56.06 
            & 0.00 & 24.77 & 57.52 
            & 0.00 & 22.91 & 55.87 
            & 0.00 & 21.73 & 54.79 \\  
            \midrule
            LongWriter-glm4-9b & Single 
            & 40.18 & 35.29 & 72.79 
            & 79.85 & 33.17 & 71.90 
            & 39.15 & 36.71 & 72.48 
            & 1.53 & 36.00 & 74.00 \\  
            \midrule
            \multirow{4}*{Qwen2.5-14B} & Single
            & 6.92 & 40.12 & 66.73     
            & 19.41 & 38.02 & 65.03    
            & 1.34 & 40.83 & 67.32     
            & 0.00 & 41.52 & 67.84 \\  
            ~ & AgentWrite
            & \underline{\bf 79.25} & 54.10 & 74.49      
            & \underline{\bf 100.00} & 51.35 & 74.22     
            & \underline{\bf 93.90} & 55.66 & 74.51      
            & \underline{\bf 43.86} & \underline{\bf 55.15} & 74.75 \\     
            ~ & Compress
            & 64.33 & 44.49 & 74.01     
            & \underline{\bf 100.00} & 43.67 & 74.87     
            & 84.84 & 44.85 & 74.24     
            & 8.14 & 44.94 & 72.93 \\   
            ~ & \agentname
            & 75.15 & \underline{\bf 55.28} & \underline{\bf 75.68}     
            & \underline{\bf 100.00} & \underline{\bf 53.17} & \underline{\bf 75.49}     
            & 87.07 & \underline{\bf 58.23} & \underline{\bf 75.93}     
            & 38.39 & 54.43 & \underline{\bf 75.62} \\  
            \midrule
            \multirow{4}*{Qwen2.5-32B} & Single
            & 4.90 & 40.46 & 66.63     
            & 14.63 & 39.06 & 66.28    
            & 0.06 & 40.69 & 66.00     
            & 0.00 & 41.63 & 67.62 \\  
            ~ & AgentWrite 
            & 61.34 & 52.39 & 74.11     
            & 97.18 & 49.98 & 73.19     
            & 74.18 & 54.08 & 74.79     
            & 12.67 & 53.10 & 74.34 \\  
            ~ & Compress
            & 52.28 & 46.47 & 73.69    
            & 92.46 & 43.31 & 73.33    
            & 60.50 & 50.17 & 74.03    
            & 3.88 & 45.94 & 73.70 \\  
            ~ & \agentname
            & \underline{\bf 77.09} & \underline{\bf 54.15} & \underline{\bf 75.67}     
            & \underline{\bf 99.83} & \underline{\bf 53.77} & \underline{\bf 74.68}     
            & \underline{\bf 91.22} & \underline{\bf 55.54} & \underline{\bf 76.83}     
            & \underline{\bf 40.22} & \underline{\bf 53.15} & \underline{\bf 75.51} \\  
            \bottomrule
        \end{tabular}
    }
    \caption{Main results on our dataset. $S_l$, $S_c$, and $S_q$ denote length, consistency, and quality scores. All reported scores are multiplied by 100. 4k, 8k, and 16k are target summary lengths. \underline{Underlined} scores are highest within each setting.}
    \label{tab:main-result}
\end{table*}

\subsection{Baselines}

\noindent \underline{\textbf{Single}} The writing instruction is fed directly to the LLM without preprocessing, and its output serves as the final result. We evaluate LongWriter-glm4-9b and GPT-4o only in this single-round setting: the former is fine-tuned for long-text generation, while the latter is restricted due to budget constraints.

\noindent \underline{\textbf{AgentWrite}~\citep{bai2024longwriter}} Similar to \agentname, AgentWrite uses structured planning and sequential paragraph composition but omits retrieval and restatement.

\noindent \underline{\textbf{Compress}} This method retains AgentWrite's workflow but preprocesses the input with LLMLingua~\cite{jiang-etal-2023-llmlingua} at a 50\% compression ratio. Its semantic-aware compression reduces the LLM's context-processing burden by discarding non-essential content while preserving key information.

\subsection{Main Results}

Table~\ref{tab:main-result} shows the main results, and we have the following observations:

\textbf{O1: More parameters do not necessarily yield stronger long-context input-output capabilities.}
Larger models do not uniformly improve performance. Under AgentWrite, Qwen2.5-32B-Instruct is lower than Qwen2.5-14B-Instruct on all three overall metrics, whereas the Single results are mixed.

\begin{figure}[t]
    \centering
    \captionsetup[subfigure]{font=scriptsize}
    \begin{minipage}[b]{0.24\columnwidth}
        \begin{subfigure}{\columnwidth}
            \centering
            \includegraphics[width=\columnwidth]{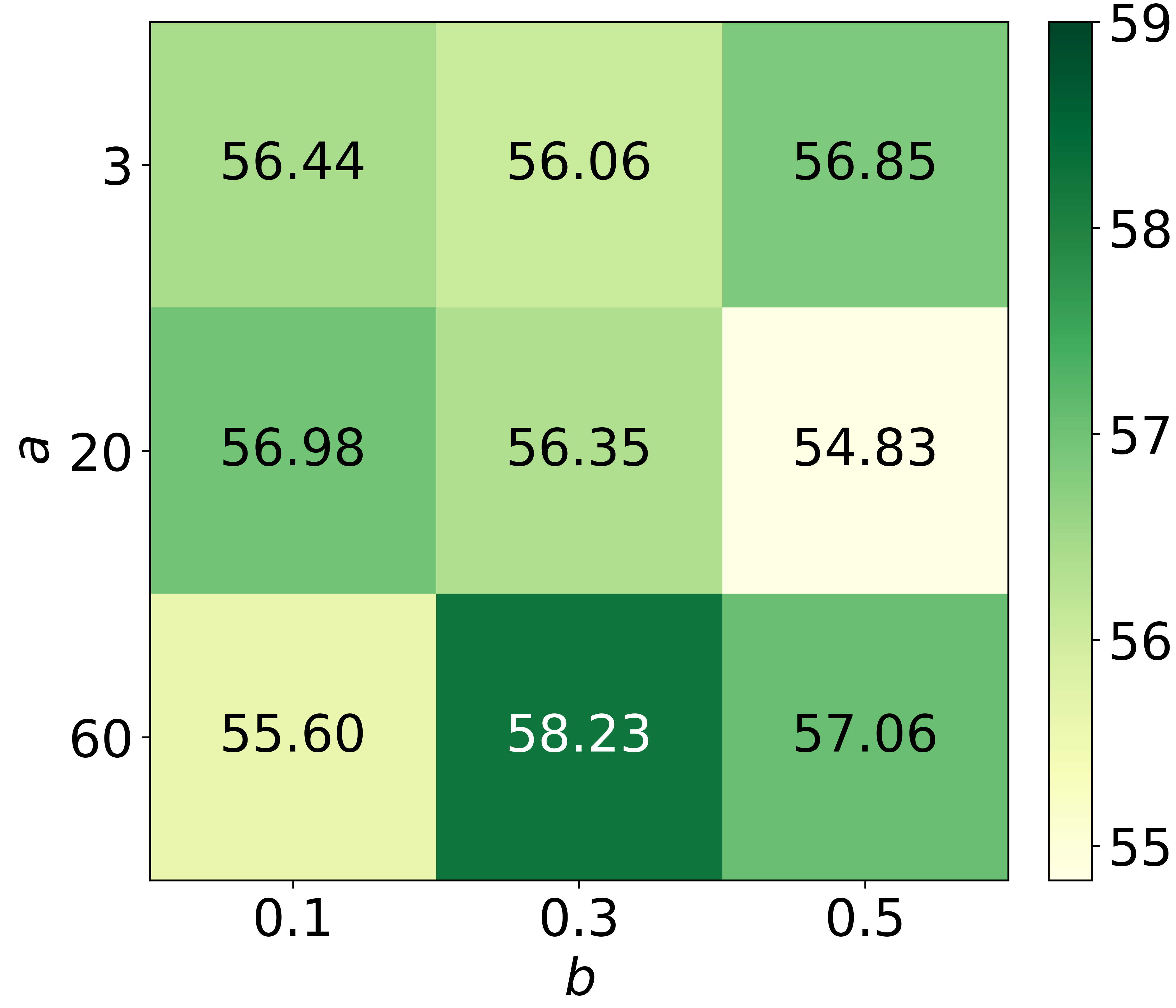}
            \caption{Heatmap of $S_c$}
            \label{fig:heat}
        \end{subfigure}
    \end{minipage}
    \begin{minipage}[b]{0.24\columnwidth}
        \begin{subfigure}{\columnwidth}
            \centering
            \includegraphics[width=\columnwidth]{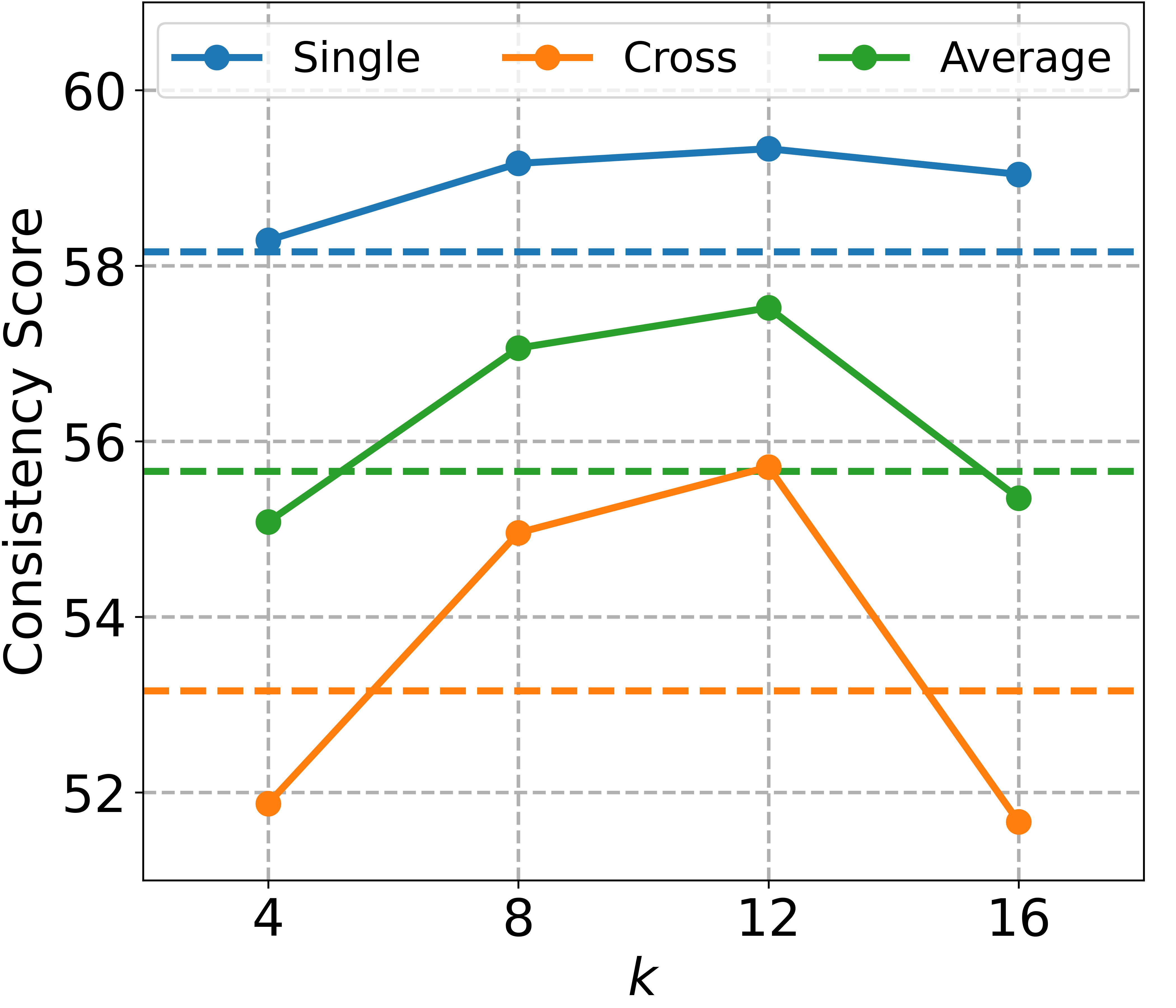}
            \caption{$S_c$ via $k$}
            \label{fig:k}
        \end{subfigure}
    \end{minipage}
    \begin{minipage}[b]{0.24\columnwidth}
        \begin{subfigure}{\columnwidth}
            \centering
            \includegraphics[width=\columnwidth]{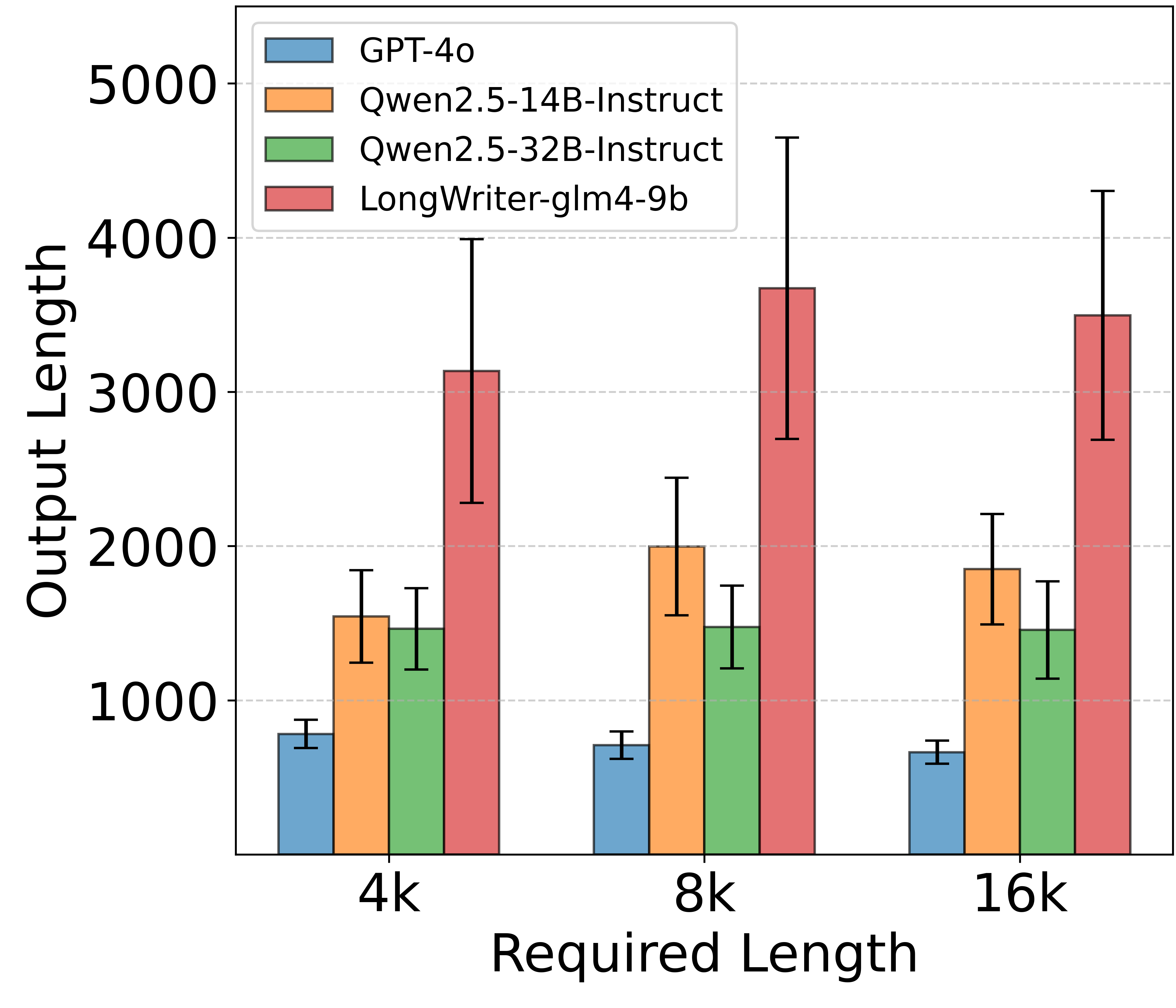}
            \caption{LLM's output length}
            \label{fig:single-length}
        \end{subfigure}
    \end{minipage}
    \begin{minipage}[b]{0.24\columnwidth}
        \begin{subfigure}{\columnwidth}
            \centering
            \includegraphics[width=\columnwidth]{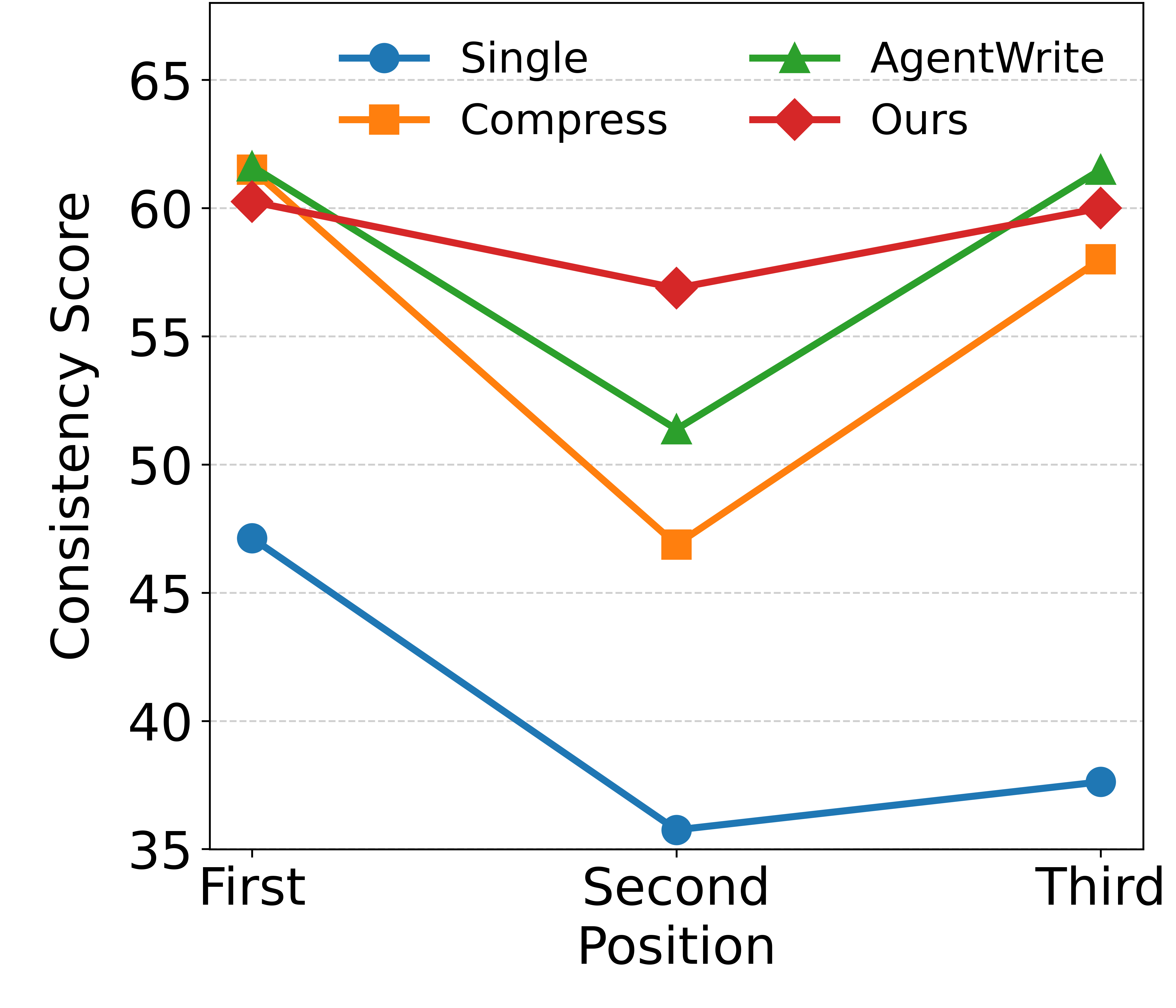}
            \caption{$\bar{S}_{\mathrm{single}}$ via Position.}
            \label{fig:pos-score}
        \end{subfigure}
    \end{minipage}
    \caption{(a) $S_c$ across $a$ and $b$. (b) $S_c$ of \agentname across $k$ values, reported for \textit{Single-Context} (``Single''), \textit{Cross-Context} (``Cross''), and their average; dashed lines indicate AgentWrite scores. (c) Output lengths of Single LLMs under varying target lengths; bars and error bars show the mean and standard deviation over 100 summaries. (d) $\bar{S}_{\mathrm{single}}$ by paper position in the input.}
\end{figure}


\textbf{O2: \agentname achieves higher consistency and quality scores while maintaining competitive length adherence.}
Across both Qwen2.5 backbones, \agentname obtains the highest overall $S_c$ and $S_q$ among the compared methods. Its overall $S_l$ is slightly lower than AgentWrite on Qwen2.5-14B-Instruct, but 15.75 points higher on Qwen2.5-32B-Instruct.

\textbf{O3: LLM agents still struggle with 16k-word generation.} Even with the Plan-Write framework, they often fail to meet the target length because some plans allocate insufficient word counts across steps. This likely reflects the \textit{Planner}'s limited ability to decompose long-context writing tasks.

\subsection{Further Analysis}

\subsubsection{Parameter Analysis} In Eq.~\eqref{eq:P_i}, $a$ and $b$ control retrieval range and strength. Evaluating all nine combinations of $a\in\{3,20,60\}$ and $b\in\{0.1,0.3,0.5\}$ using $S_c$ identifies $a=60$ and $b=0.3$ as optimal (Figure~\ref{fig:heat}). We also evaluate $k\in\{4,8,12,16\}$ with $a=10$, $b=0.2$, Qwen2.5-14B-Instruct, and an 8k-word target. Performance peaks at $k=12$ and then drops sharply (Figure~\ref{fig:k}), as larger values introduce noise and approach or exceed the 128k context limit.

\subsubsection{Analysis of Response Length} LLMs often miss target lengths in single-step generation (Table~\ref{tab:main-result}). Across 100 samples per method (Figure~\ref{fig:single-length}), GPT-4o rarely exceeds 1,000 words, Qwen2.5 averages about 2,000, and LongWriter-glm4-9b mostly remains below 4,000 words. Thus, models trained for short-input-long-output generation perform better but still miss long-input-long-output targets.


\begin{figure}[t]
    \centering
    \includegraphics[width=0.9\columnwidth]{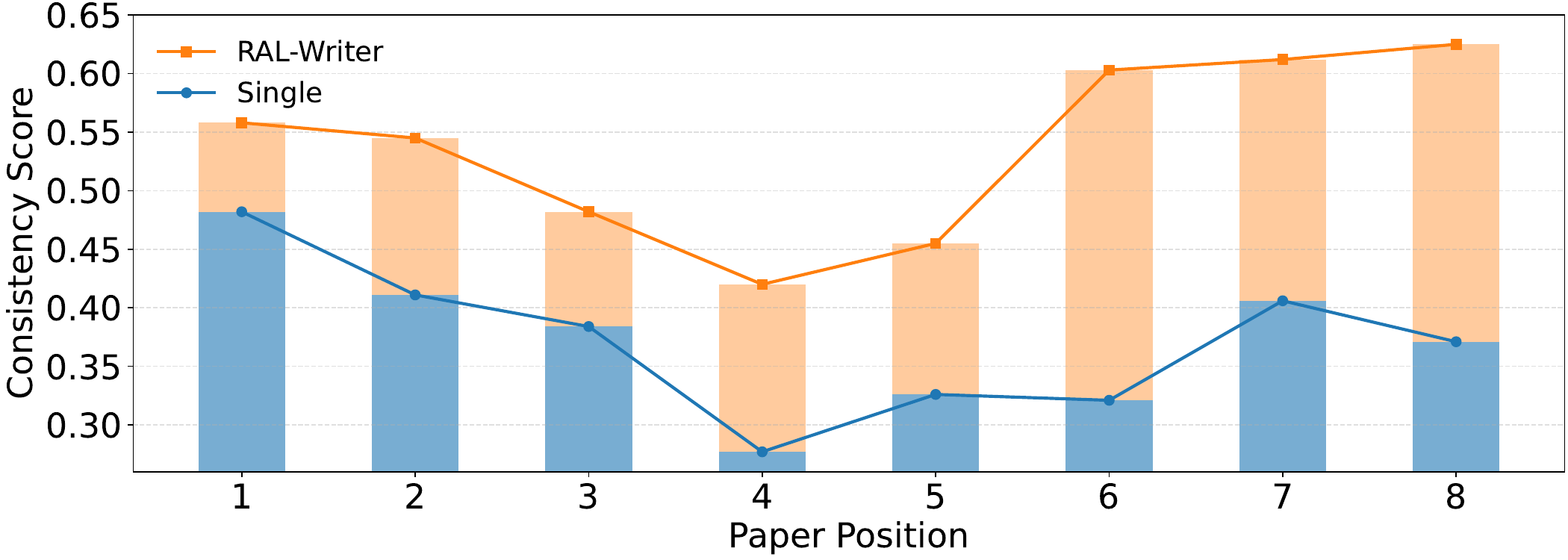}
    \caption{Consistency scores across paper positions, highlighting robustness to the lost-in-the-middle effect.}
    \label{fig:lost}
\end{figure}

\subsubsection{Position-wise Analysis}\label{subsec:lost} We verify the ``lost-in-the-middle'' effect by cyclically permuting eight papers and summarizing them with Qwen2.5-14B-Instruct. Consistency declines near the context middle (Figure~\ref{fig:lost}). With three papers, \agentname shows less middle-position degradation than the baselines (Figure~\ref{fig:pos-score}), indicating that retrieval and restatement mitigate position bias.

\section{Conclusion}


We introduce \agentname, a Plan-Write framework that effectively mitigates the ``lost-in-the-middle'' problem through targeted retrieval and iterative restatement. Our carefully curated multi-paper summarization dataset and three complementary evaluation dimensions enable a systematic study of long-input-to-long-output generation.
\section{Acknowledgement}

This work was supported by the National Science and Technology Major Project under Grant 2022ZD0120202, in part by National Natural Science Foundation of China (No. U2433212), in part by the Fundamental Research Funds for the Central Universities, and in part by the State Key Laboratory of Complex \& Critical Software Environment.

%
%
%
\bibliographystyle{splncs04}
\bibliography{refs}

@inproceedings{peng2023yarn,
  title     = {Ya{RN}: Efficient Context Window Extension of Large Language Models},
  author    = {Bowen Peng and Jeffrey Quesnelle and Honglu Fan and Enrico Shippole},
  booktitle = {The Twelfth International Conference on Learning Representations},
  year      = {2024}
}

@inproceedings{bai2024longwriter,
  title     = {LongWriter: Unleashing 10,000+ Word Generation from Long Context {LLM}s},
  author    = {Yushi Bai and Jiajie Zhang and Xin Lv and Linzhi Zheng and Siqi Zhu and Lei Hou and Yuxiao Dong and Jie Tang and Juanzi Li},
  booktitle = {The Thirteenth International Conference on Learning Representations},
  year      = {2025}
}

@inproceedings{han2024lm,
  title     = {LM-Infinite: Zero-Shot Extreme Length Generalization for Large Language Models},
  author    = {Han, Chi and Wang, Qifan and Peng, Hao and Xiong, Wenhan and Chen, Yu and Ji, Heng and Wang, Sinong},
  booktitle = {Proceedings of the 2024 Conference of the North American Chapter of the Association for Computational Linguistics: Human Language Technologies (Volume 1: Long Papers)},
  pages     = {3991--4008},
  year      = {2024}
}

@inproceedings{an2024make,
  author    = {An, Shengnan and Ma, Zexiong and Lin, Zeqi and Zheng, Nanning and Lou, Jian-Guang and Chen, Weizhu},
  booktitle = {Advances in Neural Information Processing Systems},
  pages     = {62160--62188},
  title     = {Make Your LLM Fully Utilize the Context},
  volume    = {37},
  year      = {2024}
}

@inproceedings{NEURIPS2023_91f18a12,
  author    = {Zheng, Lianmin and Chiang, Wei-Lin and Sheng, Ying and Zhuang, Siyuan and Wu, Zhanghao and Zhuang, Yonghao and Lin, Zi and Li, Zhuohan and Li, Dacheng and Xing, Eric and Zhang, Hao and Gonzalez, Joseph E and Stoica, Ion},
  booktitle = {Advances in Neural Information Processing Systems},
  pages     = {46595--46623},
  title     = {Judging LLM-as-a-Judge with MT-Bench and Chatbot Arena},
  volume    = {36},
  year      = {2023}
}

@misc{kamradt2024niah,
  title  = {Needle In A Haystack - pressure testing LLMs},
  url    = {https://github.com/gkamradt/LLMTest_NeedleInAHaystack/tree/main},
  author = {Gregory Kamradt},
  year   = {2023}
}

@inproceedings{hsieh2024ruler,
  title     = {{RULER}: What{\textquoteright}s the Real Context Size of Your Long-Context Language Models?},
  author    = {Cheng-Ping Hsieh and Simeng Sun and Samuel Kriman and Shantanu Acharya and Dima Rekesh and Fei Jia and Boris Ginsburg},
  booktitle = {First Conference on Language Modeling},
  year      = {2024}
}

@article{yang2024qwen2,
  title   = {Qwen2.5 technical report},
  author  = {Yang, An and Yang, Baosong and Zhang, Beichen and Hui, Binyuan and Zheng, Bo and Yu, Bowen and Li, Chengyuan and Liu, Dayiheng and others},
  journal = {arXiv preprint arXiv:2412.15115},
  year    = {2024}
}

@inproceedings{kwon2023efficient,
  title     = {Efficient memory management for large language model serving with pagedattention},
  author    = {Kwon, Woosuk and Li, Zhuohan and Zhuang, Siyuan and Sheng, Ying and Zheng, Lianmin and Yu, Cody Hao and Gonzalez, Joseph and Zhang, Hao and Stoica, Ion},
  booktitle = {Proceedings of the 29th Symposium on Operating Systems Principles},
  pages     = {611--626},
  year      = {2023}
}

@inproceedings{wu2024longgenbench,
  title     = {LongGenBench: Benchmarking Long-Form Generation in Long Context {LLM}s},
  author    = {Yuhao Wu and Ming Shan Hee and Zhiqiang Hu and Roy Ka-Wei Lee},
  booktitle = {The Thirteenth International Conference on Learning Representations},
  year      = {2025}
}

@article{que2024hellobench,
  title   = {Hellobench: Evaluating long text generation capabilities of large language models},
  author  = {Que, Haoran and Duan, Feiyu and He, Liqun and Mou, Yutao and Zhou, Wangchunshu and Liu, Jiaheng and Rong, Wenge and Wang, Zekun Moore and Yang, Jian and others},
  journal = {arXiv preprint arXiv:2409.16191},
  year    = {2024}
}

@article{ping2025longdpo,
  title   = {LongDPO: Unlock Better Long-form Generation Abilities for LLMs via Critique-augmented Stepwise Information},
  author  = {Ping, Bowen and Zeng, Jiali and Meng, Fandong and Wang, Shuo and Zhou, Jie and Zhang, Shanghang},
  journal = {arXiv preprint arXiv:2502.02095},
  year    = {2025}
}

@misc{Chase_LangChain_2022,
  author = {Chase, Harrison},
  title  = {LangChain},
  url    = {https://github.com/hwchase17/langchain},
  year   = {2022}
}

@inproceedings{bge_embedding,
  title     = {C-pack: Packed resources for general chinese embeddings},
  author    = {Xiao, Shitao and Liu, Zheng and Zhang, Peitian and Muennighoff, Niklas and Lian, Defu and Nie, Jian-Yun},
  booktitle = {Proceedings of the 47th international ACM SIGIR conference on research and development in information retrieval},
  pages     = {641--649},
  year      = {2024}
}

@article{quan2024language,
  title   = {Language Models can Self-Lengthen to Generate Long Texts},
  author  = {Quan, Shanghaoran and Tang, Tianyi and Yu, Bowen and Yang, An and Liu, Dayiheng and Gao, Bofei and Tu, Jianhong and Zhang, Yichang and Zhou, Jingren and Lin, Junyang},
  journal = {arXiv preprint arXiv:2410.23933},
  year    = {2024}
}

@inproceedings{he-etal-2024-never,
  title     = {Never Lost in the Middle: Mastering Long-Context Question Answering with Position-Agnostic Decompositional Training},
  author    = {He, Junqing  and
               Pan, Kunhao  and
               Dong, Xiaoqun  and
               Song, Zhuoyang  and
               Liu, YiBo  and
               Sun, Qianguo  and
               Liang, Yuxin  and
               Wang, Hao  and
               Zhang, Enming  and
               Zhang, Jiaxing},
  booktitle = {Proceedings of the 62nd Annual Meeting of the Association for Computational Linguistics (Volume 1: Long Papers)},
  year      = {2024},
  pages     = {13628--13642}
}

@inproceedings{pham-etal-2024-suri,
  title     = {{S}uri: Multi-constraint Instruction Following in Long-form Text Generation},
  author    = {Pham, Chau Minh  and
               Sun, Simeng  and
               Iyyer, Mohit},
  booktitle = {Findings of the Association for Computational Linguistics: EMNLP 2024},
  year      = {2024},
  pages     = {1722--1753}
}

@inproceedings{zhang-etal-2024-bench,
  title     = {$\infty${B}ench: Extending Long Context Evaluation Beyond 100{K} Tokens},
  author    = {Zhang, Xinrong  and
               Chen, Yingfa  and
               Hu, Shengding  and
               Xu, Zihang  and
               Chen, Junhao  and
               Hao, Moo  and
               Han, Xu  and
               Thai, Zhen  and
               Wang, Shuo  and
               Liu, Zhiyuan  and
               Sun, Maosong},
  booktitle = {Proceedings of the 62nd Annual Meeting of the Association for Computational Linguistics (Volume 1: Long Papers)},
  year      = {2024},
  pages     = {15262--15277}
}

@inproceedings{laban-etal-2024-summary,
  title     = {Summary of a Haystack: A Challenge to Long-Context {LLM}s and {RAG} Systems},
  author    = {Laban, Philippe  and
               Fabbri, Alexander  and
               Xiong, Caiming  and
               Wu, Chien-Sheng},
  booktitle = {Proceedings of the 2024 Conference on Empirical Methods in Natural Language Processing},
  year      = {2024},
  pages     = {9885--9903}
}

@inproceedings{liu-etal-2024-longgenbench,
  title     = {{L}ong{G}en{B}ench: Long-context Generation Benchmark},
  author    = {Liu, Xiang  and
               Dong, Peijie  and
               Hu, Xuming  and
               Chu, Xiaowen},
  booktitle = {Findings of the Association for Computational Linguistics: EMNLP 2024},
  year      = {2024},
  pages     = {865--883}
}

@inproceedings{tan-etal-2024-proxyqa,
  title     = {{P}roxy{QA}: An Alternative Framework for Evaluating Long-Form Text Generation with Large Language Models},
  author    = {Tan, Haochen  and
               Guo, Zhijiang  and
               Shi, Zhan  and
               Xu, Lu  and
               Liu, Zhili  and
               Feng, Yunlong  and
               Li, Xiaoguang  and
               Wang, Yasheng  and
               Shang, Lifeng  and
               Liu, Qun  and
               Song, Linqi},
  booktitle = {Proceedings of the 62nd Annual Meeting of the Association for Computational Linguistics (Volume 1: Long Papers)},
  year      = {2024},
  pages     = {6806--6827}
}

@article{liu-etal-2024-lost,
  title   = {Lost in the Middle: How Language Models Use Long Contexts},
  author  = {Liu, Nelson F.  and
             Lin, Kevin  and
             Hewitt, John  and
             Paranjape, Ashwin  and
             Bevilacqua, Michele  and
             Petroni, Fabio  and
             Liang, Percy},
  journal = {Transactions of the Association for Computational Linguistics},
  volume  = {12},
  year    = {2024},
  pages   = {157--173}
}

@inproceedings{jiang-etal-2023-llmlingua,
  title     = {{LLML}ingua: Compressing Prompts for Accelerated Inference of Large Language Models},
  author    = {Jiang, Huiqiang  and
               Wu, Qianhui  and
               Lin, Chin-Yew  and
               Yang, Yuqing  and
               Qiu, Lili},
  booktitle = {Proceedings of the 2023 Conference on Empirical Methods in Natural Language Processing},
  year      = {2023},
  pages     = {13358--13376}
}

@article{xiong2025beyond,
  title   = {Beyond Outlining: Heterogeneous Recursive Planning for Adaptive Long-form Writing with Language Models},
  author  = {Xiong, Ruibin and Chen, Yimeng and Khizbullin, Dmitrii and Zhuge, Mingchen and Schmidhuber, J{\"u}rgen},
  journal = {arXiv preprint arXiv:2503.08275},
  year    = {2025}
}

@article{wuefficient,
  title   = {An efficient recipe for long context extension via middle-focused positional encoding},
  author  = {Wu, Tong and Zhao, Yanpeng and Zheng, Zilong},
  journal = {Advances in Neural Information Processing Systems},
  volume  = {37},
  pages   = {56349--56373},
  year    = {2024}
}

@article{hurst2024gpt,
  title   = {Gpt-4o system card},
  author  = {Hurst, Aaron and Lerer, Adam and Goucher, Adam P and Perelman, Adam and Ramesh, Aditya and Clark, Aidan and Ostrow, AJ and Welihinda, Akila and Hayes, Alan and Radford, Alec and others},
  journal = {arXiv preprint arXiv:2410.21276},
  year    = {2024}
}

\end{document}